# Randomized Online CP Decomposition

Congbo Ma
School of Software Engineering
South China University of Technology
Guangzhou, P. R. China
201520121828@mail.scut.edu.cn

Xiaowei Yang
School of Software Engineering
South China University of Technology
Guangzhou, P. R. China
xwyang@scut.edu.cn

Hu Wang
Living Analytics Research Centre
Singapore Management University
Singapore, Singapore
huwang@smu.edu.sg

*Abstract*—CANDECOMP/PARAFAC (CP) decomposition has been widely used to deal with multi-way data. For real-time or large-scale tensors, based on the ideas of randomized-sampling CP decomposition algorithm and online CP decomposition algorithm, a novel CP decomposition algorithm called randomized online CP decomposition (ROCP) is proposed in this paper. The proposed algorithm can avoid forming full Khatri-Rao product, which leads to boost the speed largely and reduce memory usage. The experimental results on synthetic data and real-world data show the ROCP algorithm is able to cope with CP decomposition for large-scale tensors with arbitrary number of dimensions. In addition, ROCP can reduce the computing time and memory usage dramatically, especially for large-scale tensors.

*Keywords—CP Decomposition; Tensor Decomposition; Randomized-Sampling; Online Learning*

## I. INTRODUCTION

CANDECOMP/PARAFAC (CP) decomposition is an important technique for data mining, dimensionality reduction, pattern recognition, object detection, classification, gene clustering, sparse representation and coding. It has had wide applications such as graph and network analysis [3], blind source separation [4], neuroscience [5,6], signal processing [7], computer vision [8]. In these applications, CP decomposition is always computed via off-line optimization algorithms [9]. However, in many applications, the data are serially acquired from users and can be transferred into a large-scale tensor. In these cases, they cannot be globally modeled by a low-rank CP model. Therefore, novel online tensor decomposition algorithms are needed to deal with such dynamic tensors.

In the previous studies, most of the existing CP decomposition algorithms, especially the alternating least squares (ALS) algorithm [1], are off-line algorithms and need to compute Khatri–Rao products of tall factors and multiplication of large matrices, which requires high computational cost and large memory. Therefore, the previous off-line CP decomposition algorithms are not suitable for very large-scale tensors. In order to deal with CP decomposition for large-scale tensor, based on the fact that the randomized methods have been used successfully for solving linear least squares problems [17,18], Battaglino et al. [2] proposed a randomized CP-ALS. To the best of our knowledge, it is an off-line algorithm and cannot be directly used for online tensor decomposition. Based on the assumption that the observed tensor at time t+1 is obtained from that at time t after appending a new slice in the time dimension, Nion and Sidiropoulos [10] proposed simultaneous diagonalization tracking (PARAFAC-SDT) algorithm and recursive least squares tracking (PARAFAC-RLST) algorithm to track the online CP decomposition of a third-order tensor. Unfortunately, the time consuming of two algorithms depends on the cost of SVD, which has high complexity, and limits their applications on large-scale tensor. To avoid computing Khatri–Rao products, Phan [11] divided a large-scale tensor into a grid of multiple subtensors and proposed a grid CP decomposition algorithm for tensor with arbitrary dimensions, in which only Hadamard products, which are multiplication of small matrices, are calculated. Recently, Zhou [12] proposed an efficient online CP decomposition algorithm that can incrementally track the CP decompositions of dynamic tensors with an arbitrary number of dimensions. This algorithm can efficiently track the new decomposition by using complementary matrices to temporally store the useful information of the previous time step. The main reason is that it makes good use of the intermediate results and avoids duplicated operations.

Motivated by the ideas in [2] and [12], in this study, we propose an efficient CP decomposition method based on randomized-sampling to deal with decomposition problem of dynamic large-scale tensors. In the proposed algorithm, randomized sampling strategy is used to form the reduced Khatri-Rao product matrices and unfolding matrices, which leads that the computing time and memory usage have been reduced largely, especially for the large-scale tensor data.

The rest of this paper is organized as follows: Section 2 first gives tensor notations and basic operations, and then briefly reviews the CP decomposition based on random sampling. Section 3 presents the proposed randomized online CP decomposition algorithm. The evaluation of the computational performance of the proposed algorithm is reported in Section 4. Finally, Section 5 makes a conclusion and outlines further directions.

## II. PRELIMINARIES

### A. Tensor Notations and Basic Operations

**Definition 1 (Tensor):** Tensors are considered as multidimensional arrays in data analysis [1]. The order of a tensor refers the number of modes. A tensor here is denoted by Euler script capital letter, e.g. $\chi \in \mathbb{R}^{I_1 \times \cdots \times I_N}$. Tensor entry $x_{i_1,i_2,\ldots,i_N}$ is mapped to entry $(i_n, j)$ of $X_{(n)}$ via the relation [2]:

$$j = 1 + \sum_{\substack{k=1 \\ k \neq n}}^{N}(i_k - 1)J_k \quad \text{where } J_k = \prod_{\substack{m=1 \\ m \neq n}}^{k=1} I_m \quad (1)$$

Where the transpose, inverse, and pseudo-inverse of a matrix are denoted by $U^T$, $U^{-1}$ and $U^{\dagger}$, respectively.

**Definition 2 (Frobenius Norm):** For a tensor $\chi \in \mathbb{R}^{I_1 \times \cdots \times I_N}$, the Frobenius Norm is the square root of the sum of the squares of all elements:

$$\|\chi\|_F = \sqrt{\sum_{i_1=1}^{I_1}\sum_{i_2=1}^{I_2}\cdots\sum_{i_N=1}^{I_N} x_{i_1,i_2,\ldots,i_N}^2} \quad (2)$$

**Definition 3 (Khatri-Rao Product):** For any of the two matrices $A \in \mathbb{R}^{M \times N}$, $B \in \mathbb{R}^{P \times Q}$, their Kronecker product is

$$A \otimes B = \begin{bmatrix} a_{11}B & \cdots & a_{1N}B \\ \vdots & \ddots & \vdots \\ a_{M1}B & \cdots & a_{MN}B \end{bmatrix} \in \mathbb{R}^{MP \times NQ} \quad (3)$$

For $N = Q$, their Khatri-Rao product is

$$A \odot B = [a_1 \otimes b_1, a_2 \otimes b_2, \ldots, a_N \otimes b_N] \quad (4)$$

Where $\odot_{i \neq n}^{N} U^{(i)}$ denotes the Khatri-Rao product of a series of loading matrices $U^{(1)}, U^{(2)}, \ldots, U^{(N)}$.

**Definition 4 (Hadamard Product):** For two same sizes matrices $A \in \mathbb{R}^{M \times N}, B \in \mathbb{R}^{M \times N}$, their Hadamard product is

$$A \circledast B = \begin{bmatrix} a_{11}b_{11} & \cdots & a_{n1}b_{n1} \\ \vdots & \ddots & \vdots \\ a_{M1}b_{M1} & \cdots & a_{MN}b_{MN} \end{bmatrix} \in \mathbb{R}^{MN} \quad (5)$$

Where $\circledast_{i \neq n}^{N} U^{(i)}$ denotes the Hadamard product of a series of loading matrices $U^{(1)}, U^{(2)}, \ldots, U^{(N)}$.

**Definition 5 (Alternating Least Squares of CP Decomposition):** The standard method for CP decomposition is alternating least squares [1], abbreviated as CP-ALS. Basically, for an $N$-order tensor $\chi \in \mathbb{R}^{I_1 \times \cdots \times I_N}$, the objective is to solve the linear least squares problem given by

$$\underset{U^{(n)}}{\arg\min} \left\| X_{(n)} - U^{(n)} Z_{(n)}^T \right\|_F \quad (6)$$

Where $U^{(n)} \in \mathbb{R}^{I_n \times R}$ is a loading matrix, $X_{(n)} \in \mathbb{R}^{I_n \times \prod_{i \neq n}^{N} I_i}$ is mode-n metricized version of a tensor $\chi$, $Z_{(n)} \in \mathbb{R}^{\prod_{i \neq n}^{N} I_i \times R}$ is Khatri-Rao product, and R is the tensor rank. CP-ALS fixes all loading matrices but $U^{(n)}$ to solve the optimization problem in (6).

### B. CP Decomposition Based on Random Sampling

Based on the fact that the randomized methods have been used successfully for solving linear least squares problems [17,18], Battaglino et al. [2] proposed a randomized version of CP-ALS called CPRAND to solve a sampled version of the least squares problem in (6). Obviously, CPRAND is an off-line randomized CP decomposition algorithm based on CP-ALS. Based on Eq. (6), we consider the following optimization problem:

$$\underset{U^{(n)}}{\arg\min} \left\| Z_{(n)}(U^{(n)})^T - (X^{(n)})^T \right\|_F \quad (7)$$

For a given number s which stands for the number of sampling, the sampling algorithm samples $s$ rows from $(X_{(n)})^T$ and the corresponding rows from $Z_{(n)}$. Then, $Z_{s(n)}$ and $(X_{s(n)})^T$ is used to denote $Z_{(n)}$ and $(X_{(n)})^T$ after sampling [2].

After sampling the $j$-th row of $Z_{(n)}$, using the mapping in Eq. (1), we can get a serial of indices $(i_1, \ldots, i_{n-1}, i_{n+1}, \ldots, i_N)$ from $j$. The $j$-th row of $Z_{(n)}$ is the Hadamard products of the corresponding rows of loading matrices, i.e.

$$Z_{s(n)}(j,:) = U^{(1)}(i_1,:) \circledast \ldots \circledast U^{(n-1)}(i_{n-1},:) \circledast U^{(n+1)}(i_{n+1},:) \\ \circledast \ldots \circledast U^{(N)}(i_N,:) \quad (8)$$

Which means CPRAND uses Hadamard products instead of Khatri–Rao products to calculate $Z_{s(n)}$. By doing this, it can reduce processing time greatly as well as the allocated memory.

So the problem in (7) could be transferred into the following problem based on the random sampling.

$$\underset{U^{(n)}}{\arg\min} \left\| Z_{s(n)}(U^{(n)})^T - (X_{s(n)})^T \right\|_F \quad (9)$$

Where the size of sampled unfolding matrix $(X_{s(n)})^T$ is $s \times I_n$, the size of sampled Khatri-Rao matrix $Z_{s(n)}$ is $s \times R$, and the solution $(U^{(n)})^T$ is of size $R \times I_n$.

## III. RANDOMIZED ONLINE CP DECOMPOSITION

In order to deal with CP decomposition for online tensor, motivated by the idea in [2], we propose a randomized online CP decomposition algorithm (ROCP) in this part. For convenience, the initial tensor is denoted by $\chi \in \mathbb{R}^{I_1 \times \cdots \times I_{N-1} \times I_{N(old)}}$, the newly coming tensor is denoted by $\chi_{new} \in \mathbb{R}^{I_1 \times \cdots \times I_{N-1} \times I_{N(new)}}$, where $I_{N(old)} \gg I_{N(new)}$. For online tensor, only the last mode $I_N$ changed and other modes $I_1, \ldots, I_{N-1}$ keep unchanged. In the following, we firstly give the updating formulations of factor matrices and then give the initialization description of ROCP.

### A. Update Steps of Randomized Online CP decomposition

During running the ROCP algorithm, on the one hand, we need to update the last factor matrix $U^{(N)}$, on the other hand, we need to update the other factor matrices $U^{(n)}$. In the following, we will give the detailed updating formulations.

*1) Update the last mode $U^{(N)}$*

From the initial step, we can obtain $(X_{s(N)old})^T$. When a new slice of tensor comes, $(X_{s(N)})^T$ is divided into two parts: $(X_{s(N)old})^T$ and $(X_{s(N)new})^T$. $(X_{s(N)new})^T$ is the sampled unfolding matrix of newly coming tensor $\chi_{new} \in \mathbb{R}^{I_1 \times \cdots \times I_{n-1} \times I_{N(new)}}$. From Eq. (9), $U^{(N)}$ can be expressed by:

$$U^{(N)} \leftarrow \underset{U^{(N)}}{\arg\min} \left\| Z_{s(N)}(U^{(N)})^T - (X_{s(N)})^T \right\|_F$$

$$= \underset{U^{(N)}}{\arg\min} \left\| Z_{s(N)} \begin{bmatrix} U_1^{(N)} \\ U_2^{(N)} \end{bmatrix}^T - \begin{bmatrix} X_{s(N)old} \\ X_{s(N)new} \end{bmatrix}^T \right\|_F$$

$$= \underset{U^{(N)}}{\arg\min} \left\| \begin{bmatrix} Z_{s(N)}(U_1^{(N)})^T - (X_{s(N)old})^T \\ Z_{s(N)}(U_2^{(N)})^T - (X_{s(N)new})^T \end{bmatrix} \right\|_F \quad (10)$$

Where $Z_{s(N)} = \circledast^{N-1} U_s^{(i)}$. Calculating directly full Khatri-Rao matrix and least squares is expensive. To improve the efficiency, we sample $Z_{(N)}$ and $X_{(N)}$. $X_{s(N)old}$ is of size $I_N \times s$. To merge $X_{s(N)old}$ and $X_{s(N)new}$ into $\begin{bmatrix} X_{s(N)old} \\ X_{s(N)new} \end{bmatrix}$, the column number of $X_{s(N)old}$ and $X_{s(N)new}$ should be the same. The size of $X_{s(N)new}$ should be $I_{N(new)} \times s$, where $s$ columns are selected out of $X_{(N)new}$ to form $X_{s(N)new}$. In order to ensure the randomness of sampling, the indexes of selected columns of $X_{(N)new}$ are not the same as $X_{(N)old}$, which lead to that $X_{(N)new}$ will be randomly sampled each time. This is illustrated in Figure 1.

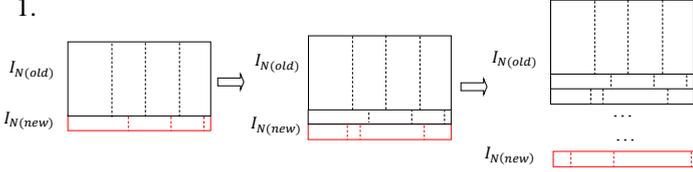

Fig. 1. Randomly sampled columns from $X_{(N)new}$

Next, we sample $Z_{(N)}$ to form $Z_{s(N)}$, which is calculated from $U^{(1)}, U^{(2)}, \ldots, U^{(N-1)}$. The indexes of selected rows of $Z_{(N)}$ should be the same indexes when forming $(X_{s(N)new})^T$. Previously calculated loading matrix $U^{(1)} \ldots U^{(N-1)}$ are used to form $Z_{s(N)}$.

The new loading matrix $U^{(N)}$ consists of two parts:

$$U^{(N)} = \begin{bmatrix} U_{old}^{(N)} \\ U_{new}^{(N)} \end{bmatrix} = \begin{bmatrix} U_{old}^{(N)} \\ X_{s(N)new}((\odot^{N-1} U_s^{(i)})^T)^\dagger \end{bmatrix} = \begin{bmatrix} U_{old}^{(N)} \\ X_{s(N)new}((Z_{s(n)})^T)^\dagger \end{bmatrix} \quad (11)$$

The upper part is previously calculated loading matrix $U_{old}^{(N)}$ and the lower part is the result from least squares of a slice of newly coming data $\chi_{new} \in \mathbb{R}^{I_1 \times \cdots \times I_{n-1} \times I_{N(new)}}$ after sampling.

*2) Update other modes $U^{(n)}$*

The loading matrices of other modes can be updated based on the following optimization problem

$$U^{(n)} \leftarrow \underset{U^{(n)}}{\arg\min} \left\| Z_{s(n)}(U^{(n)})^T - (X_{s(n)})^T \right\|_F \quad (12)$$

Where $Z_{s(n)} = U_s^{(N)} \circledast \ldots \circledast U_s^{(n+1)} \circledast U_s^{(n-1)} \circledast \ldots \circledast U_s^{(1)}$. The results from least squares are

$$U^{(n)} = X_{s(n)} Z_{s(n)} \left( (Z_{s(n)})^T Z_{s(n)} \right)^\dagger = PQ^{-1} \quad (13)$$

Where $P = X_{s(n)} Z_{s(n)}$ and $Q = (Z_{s(n)})^T Z_{s(n)}$. In order to simplify the calculation, the calculation of $U^{(n)}$ is divided into two parts: $P$ and $Q$.

When the new slice of tensor comes, $X_{s(n)}$ can be divided into two parts: $X_{s(n)old}$ and $X_{s(n)new}$ and $P$ can be expressed as

$$P = X_{s(n)} Z_{s(n)}$$

$$= [X_{s(n)old}, X_{s(n)new}] \left( \begin{bmatrix} U_{old}^{(N)} \\ U_{new}^{(N)} \end{bmatrix} \circledast \ldots \circledast U_s^{(n+1)} \circledast U_s^{(n-1)} \circledast \ldots \circledast U_s^{(1)} \right)$$

$$= [X_{s(n)old}, X_{s(n)new}] \begin{bmatrix} U_{old}^{(N)} \circledast \ldots \circledast U_s^{(n+1)} \circledast U_s^{(n-1)} \circledast \ldots \circledast U_s^{(1)} \\ U_{new}^{(N)} \circledast \ldots \circledast U_s^{(n+1)} \circledast U_s^{(n-1)} \circledast \ldots \circledast U_s^{(1)} \end{bmatrix}$$

$$= X_{s(n)old} \left( U_{old}^{(N)} \circledast \ldots \circledast U_s^{(n+1)} \circledast U_s^{(n-1)} \circledast \ldots \circledast U_s^{(1)} \right)$$

$$+ X_{s(n)new} \left( U_{new}^{(N)} \circledast \ldots \circledast U_s^{(n+1)} \circledast U_s^{(n-1)} \circledast \ldots \circledast U_s^{(1)} \right)$$

$$= P_{old} + X_{s(n)new} (U_{new}^{(N)} \circledast \ldots \circledast U_s^{(n+1)} \circledast U_s^{(n-1)} \circledast \ldots \circledast U_s^{(1)})$$

$$= P_{old} + X_{s(n)new} Z_{s(n)}^{new} \quad (14)$$

Where $P_{old}$ is the result of the previous step and $P$ can be updated just from the calculation of $X_{s(n)new}$ and $Z_{s(n)}^{new}$. $s$ columns have been selected from $X_{(n)}$ at the initial step, so $s$ rows should be selected from $(X_{s(n)new})^T$ and $Z_{s(n)}^{new}$. This is because we used Hadamard products of loading matrices to get $Z_{s(n)}^{new}$, and the definition of Hadamard products demands the same rows and columns of matrices. Therefore $U_{new}^{(N)}$ and other loading matrices $U_s^{(i)}$ have the same size of $s \times R$. Obviously, the size of $Z_{s(n)}^{new}$ is $s \times R$. The indexes of selected rows of $Z_{s(n)}^{new}$ is the same indexes of as $(X_{s(n)new})^T$. $Z_{s(n)}^{new}$ is calculated by

$$Z_{s(n)}^{new} = SKR\left(s, U^{(1)} \ldots, U^{(n-1)}, U^{(n+1)}, \ldots, U^{(N-1)}, U_{new}^{(N)}\right) \quad (15)$$

The Sampled Khatri-Rao Product (SKR) algorithm is shown in Algorithm 1, in which $j$ are indexes of sampling rows and $\{i_1^{(j)}, \ldots, i_{n-1}^{(j)}, i_{n+1}^{(j)}, \ldots, i_N^{(j)}\}$, which can be calculated from Eq. (1), is a set of indexes of the corresponding loading matrices [2].

| Algorithm 1: Sampled Khatri-Rao Product |
|---|
| **Input:** Number of sampling: $s$ |
|        Loading matrices: $U^{(1)}, \ldots, U^{(n-1)}, U^{(n+1)} \ldots, U^{(N)}$ |
| **Output:** Khatri-Rao product after sampling: $Z_{s(n)}$ |
| 1    Retrieve idxs ← $\{i_1^{(j)}, \ldots, i_{n-1}^{(j)}, i_{n+1}^{(j)}, \ldots, i_N^{(j)}\}$ |
| 2    $Z_{s(n)} \leftarrow 1$ |
| 3    for m=1,…,n-1,n+1,…,N |
| 4        $U_s^{(m)} \leftarrow U^{(m)}$(idxs(:,m),:)      % Matlab style code |
| 5        $Z_{s(n)} \leftarrow Z_{s(n)} \circledast U_s^{(m)}$ |
| 6    end for |

Similarly, $Q$ can be calculated by

$$Q = Q_{old} + (Z_{s(n)}^{new})^T Z_{s(n)}^{new} \quad (16)$$

Thus, based on the complementary matrices $P$, $Q$, the loading matrix $U^{(n)}$ can be calculated by the following equation

$$U^{(n)} = PQ^{-1} \quad (17)$$

Based on the above analysis, the update procedure of the proposed algorithm is given in Algorithm 2.

---

**Algorithm 2:** Randomized-Sampling Online CP Decomposition Update

---

**Input:** The new incoming data tensor: $\chi_{new}$
  Previous loading matrices of $\chi_{init}$: $U^{(1)}, U^{(2)}, …, U^{(N-1)}$
  Complementary matrices: $P^{(1)}, P^{(2)}, …, P^{(N-1)}$ and $Q^{(1)}, Q^{(2)}, …, Q^{(N-1)}$
**Output:** Updated loading matrices: $U^{(1)}, U^{(2)}, …, U^{(N)}$
1  Define sampling operator $s$
   //Update The Last Mode $U^{(N)}$
2  $X_{s(N)new} \leftarrow$ sampling $(X_N, s)$
3  $Z_{s(N)} \leftarrow SKR(S, U^{(1)}, U^{(2)}, …, U^{(N-1)})$
4  $V \leftarrow (Z_{s(N)})^T Z_{s(N)}$
5  $U_{new}^{(N)} \leftarrow X_{s(N)new} Z_{s(N)} V^{-1}$
6  $U^{(N)} \leftarrow \begin{bmatrix} U_{old}^{(N)} \\ U_{new}^{(N)} \end{bmatrix}$
   // Update The Other Modes
7  **for** n = 1:N-1 **do**
8    $X_{s(n)new} \leftarrow$ sampling $(X_n, s)$
9    $Z_{s(n)}^{new} \leftarrow SKR(S, U^{(1)},…, U^{(n-1)}, U^{(n+1)},…,U^{(N-1)}, U_{new}^{(N)})$
10   $P^{(n)} \leftarrow P^{(n)} + X_{s(n)new} Z_{s(n)}^{new}$
11   $Q^{(n)} \leftarrow Q^{(n)} + (Z_{s(n)}^{new})^T Z_{s(n)}^{new}$
12   $U^{(n)} \leftarrow P^{(n)} (Q^{(n)})^{-1}$
13 **end for**

---

### B. Initialization of ROCP

In the initialization step of ROCP, we need to give complementary matrices $P$ and $Q$. In the process of iterations of initial step, we need to store the sampled Khatri-Rao matrices $Zs$ and sampled Khatri-Rao matrices $U$ with the highest fitness, which are denoted by $Zs\_best$ and $U\_best$, respectively. The detailed initialization procedure of ROCP is given in Algorithm 3.

---

**Algorithm 3:** The Initialization Step of ROCP

---

**Input:** Initial tensor: $\chi_{init}$
  Best initial Khatri-Rao Product after sampling: $Zs\_best^{(1)}, Zs\_best^{(2)},…,Zs\_best^{(N-1)}$
  Best initial loading after sampling: $U\_best^{(1)}, U\_best^{(2)},…, U\_best^{(N-1)}$
**Output:** Complementary matrices: $P^{(1)}, P^{(2)}, …, P^{(N-1)}$ and $Q^{(1)}, Q^{(2)},…, Q^{(N-1)}$
1  Get $Zs\_best$ and $U\_best$ from Sampling
2  **for** n= 1:N-1 **do**
3    $P^{(n)} \leftarrow U\_best^{(n)}$
4    $Q^{(n)} \leftarrow (Zs\_best^{(n)})^T * Zs\_best^{(n)}$
5  **end for**

---

Based on the above analysis, the complete ROCP algorithm is shown.

---

**Algorithm 4:** ROCP Algorithm

---

**Input:** Initial tensor: $\chi_{init}$
  The new incoming data tensor: $\chi_{new(1)}, \chi_{new(2)}, …, \chi_{new(k)}$
  Number of sampling: $s$
**Output:** Loading matrices: $U^{(1)}, U^{(2)}, …, U^{(N)}$
1  Using CP decomposition based on Random Sampling [2] to decompose $\chi_{init}$
2  Using **Algorithm 3** to initialize ROCP
3  **for** i= 1:k **do**
4    Using **Algorithm 2** to update loading matrices: $U^{(1)}, U^{(2)}, …, U^{(N)}$
5  **end for**

---

## IV. EXPERIMENTS

In order to empirically evaluate the effectiveness and efficiency of ROCP algorithm, we conduct some experiments on six different structure synthetic datasets and six real datasets, and compare it with three existing state-of-the-art methods. In the following, we introduce datasets used in our analysis, describe the parameter settings, and give the experimental results.

### A. Experimental setting

All experiments are performed on MATLAB R2015b using Tensor Toolbox v2.6 [13] on an Intel Xeon E5-2620 v2 2.1 GHz machine with 64 GB of memory. In all of the experiments, 20 percent of the total data is used as the initial tensor, and the remaining 80 percent data serve as dynamic data newly coming each time. 10 trials have been run for each experiment and the average results of these 10 trials have been reported.

Batch Cold, Batch Hot, OnlineCP [12] are selected as baselines to evaluate the performance of the proposed algorithm. Batch Cold and Batch Hot are off-line CP-ALS algorithms with different initialization. These three methods have been listed as below:

- Batch Cold, we directly use a Matlab version of CP-ALS implementation in open-source Tensor Toolbox [13]. Here, we simply recomputed batch Cold each time when a new slice of data arrives.

- Batch Hot in which the result of the CP decomposition from the previous step is used as the initialization of CP decomposition on current step.

- OnlineCP method is an online CP decomposition method that is the most recent which is related to our work. OnlineCP updates so-called temporal mode and non-temporal mode separately by using information from the previous step. It also uses a dynamic programming strategy to compute all Khatri-Rao product in one run by making good use of intermediate results and avoiding duplicated operations.

In our experiments, PARAFAC-SDT, PARAFAC-RLST [10] and grid CP decomposition [11] have not been compared with the proposed algorithm. The reason is that on the one hand, OnlineCP is much better than these three algorithms in terms of computational time and approximate accuracy [12]. On the other hand, PARAFAC-SDT and PARAFAC-RLST can only handle three order tensors and cannot deal with higher order tensors.

There are some settings of parameters which need to be clarified. For the initialization of ROCP, the tolerance is 1e-4 for synthetic datasets and 1e-3 for real datasets, the maximum number of iterations is 100 for all datasets. In ROCP, the sample size $s$ is 10Rlog(R) [2] in the initial step and online process. In term of the specific setting of other benchmarks, for Batch Cold and Batch Hot algorithms, the default tolerance and the maximum number of iterations are 1e-4 and 50, respectively. For OnlineCP, we use the default best parameters in [12]: the tolerance and the maximum number of iterations are set to 1e-8 and 100, respectively.

Fitness and average running time are two performance metrics and the definition of fitness is given as below:

$$fitness = \left(1 - \frac{\|\hat{x} - \chi\|}{\|\chi\|}\right)$$

Where $\chi$ is the original tensor and $\hat{x}$ is the tensor after decomposition operation.

### B. Synthetic Datasets

Six synthetic data are constructed from random loading matrices, which are downgraded by a Gaussian noise with a Signal-to-Interference Rate (SIR) of 20 dB [12]. The rank R is fixed to 5 for all tensors. The computational time of four methods is tested with different orders and different sizes of tensors. The average time is reported in Table I.

Table I shows that ROCP is faster than OnlineCP, Batch Cold, Batch Hot. Especially, ROCP has achieved higher speedup ratio for larger data. For example, the speedup ratio of ROCP with respect to OnlineCP is 1.87 on the data of dimensions of [400*400*200] while the speedup ratio of ROCP with respect to OnlineCP reaches 6.40 on the data of dimensions of [20*20*20*20*200]. The reason is that ROCP uses random sampling to avoid explicitly forming full Khatri-Rao product and unfolding matrices.

TABLE I. AVERAGE RUNNING TIME (SECONDS)

| Dimensions | Batch Cold | Batch Hot | OnlineCP | ROCP |
|---|---|---|---|---|
| 400*400*200 | 249.1174 | 93.4628 | 2.3720 | **1.2670** |
| 500*500*200 | 356.4676 | 137.7290 | 3.2798 | **1.5826** |
| 60*60*60*200 | 493.1831 | 159.9477 | 3.6174 | **1.8408** |
| 100*100*100*200 | 1391.2022 | 593.1684 | 15.9311 | **5.6266** |
| 20*20*20*20*200 | 507.0369 | 131.0944 | 11.6727 | **1.8232** |
| 30*30*30*30*200 | 1564.0085 | 486.1832 | 20.1512 | **4.3981** |

### C. Real Datasets

Table II lists six real datasets used in our experiments. All of these datasets are surveillance videos which are natural four-way structures. The first three-way data of video are images of each frame and the fourth-way is time dimension. The contents of videos are listed as follows: (i) CWSi: people with helmet in construction working sites, where several people work in construction sites with helmet, while others do not. (ii) Camera1 and Camera2: videos that contain suspicious human actions, several people walk back and forth through the sight of cameras. (iii) Indoor: videos shot in indoor environments, which contains people sitting in chairs and talk. (v) Outdoor: videos shot in outdoor environments, three people and three cars act on a robbery sequence, where suitcases and bags are carried, left and picked from the floor. (vi) Seq1: Two people walk towards each other and after some unknown event one runs away and the other tails him. In term of settings, if the fourth-way tensor is larger, the batch size of ROCP will be bigger correspondingly. If the first two ways of data are large, we tune R larger correspondingly. The experimental results are given in Tables III-IV.

TABLE II. CHARACTERISTICS OF REAL DATASETS

| Name | Dimensions | SliceSize $S = \prod_{i=1}^{N-1} I_i$ | Batch size | R |
|---|---|---|---|---|
| CWSi[15] | 600*800*3*31 | 1,440,000 | 1 | 5 |
| Camera1[15] | 288*384*3*500 | 442,368 | 10 | 5 |
| Camera2[15] | 288*384*3*500 | 442,368 | 10 | 5 |
| Indoor[16] | 1,040*1,392*3*100 | 4,343,040 | 10 | 6 |
| Outdoor[16] | 1,040*1,392*3*100 | 4,343,040 | 10 | 7 |
| Seq1[14] | 480*640*3*221 | 921,600 | 10 | 5 |

TABLE III. AVERAGE FITNESS OF FOUR ALGORITHMS ON REAL DATASETS. FOR ONLINECP AND ROCP, THE RATIOS OF THEIR FITNESS WITH RESPECTIVE TO BATCH HOT ARE SHOWN IN BRACKETS. BOLDFACE INDICATES THE BEST RESULTS.

| Datasets | Batch Cold | Batch Hot | OnlineCP | ROCP |
|---|---|---|---|---|
| CWSi | 0.7570 | 0.7625 | **0.7351(0.96)** | 0.7062(0.92) |
| Camera1 | 0.7926 | 0.7927 | **0.7926(0.99)** | 0.7459(0.94) |
| Camera2 | 0.8244 | 0.8246 | **0.8239(0.99)** | 0.7910(0.96) |
| Indoor | 0.8506 | 0.8515 | **0.8506(0.99)** | 0.8329(0.98) |
| Outdoor | 0.8032 | 0.8039 | **0.8034(0.99)** | 0.7803(0.97) |
| Seq1 | 0.7961 | 0.7964 | **0.7959(0.99)** | 0.7639(0.96) |

TABLE IV. AVERAGE RUNNING TIME OF FOUR ALGORITHMS ON REAL DATASETS. FOR ONLINECP AND ROCP, THE SPEEDUP RATIOS OF THE RUNNING TIME WITH RESPECTIVE TO BATCH HOT ARE SHOWN IN BRACKETS. BOLDFACE INDICATES THE BEST RESULTS.

| Datasets | Batch Cold | Batch Hot | OnlineCP | ROCP |
|---|---|---|---|---|
| CWSi | 253.5743 | 75.1817 | 4.6345(16.22) | **1.7088(44.00)** |
| Camera1 | 561.6725 | 114.8558 | 7.0959(16.19) | **3.1998(35.89)** |
| Camera2 | 643.8407 | 141.6831 | 8.7273(16.24) | **4.5271(31.30)** |
| Indoor | 471.3334 | 56.4390 | 17.9202(3.15) | **7.1893(7.85)** |
| Outdoor | 358.8775 | 63.4169 | 19.9781(3.17) | **7.4998(8.46)** |
| Seq1 | 295.7805 | 69.6882 | 7.8740(8.85) | **3.5319(19.73)** |

From Tables III and IV, we can find that the two off-line CP decomposition algorithms are still very time consuming, on-line algorithms have demonstrated overwhelming advantages on speed. The proposed ROCP algorithm achieves the fastest executive speed and good fitness in real datasets.

### D. Scalability of ROCP

To confirm the scalability of ROCP, we design following two experiments:

*1) A tensor $\chi \in \mathbb{R}^{60 \times 60 \times 60 \times 1000}$ with large time dimension will be decomposed. The batch size of a newly incoming slice of tensor each time is 1, and we record the running time through time units.*

*2) A tensor $\chi \in \mathbb{R}^{60 \times 60 \times 60 \times 200}$ will be decomposed. The batch size of a newly incoming slice of tensor each time ranges from 1 to 10, and we record the running time through different batch sizes.*

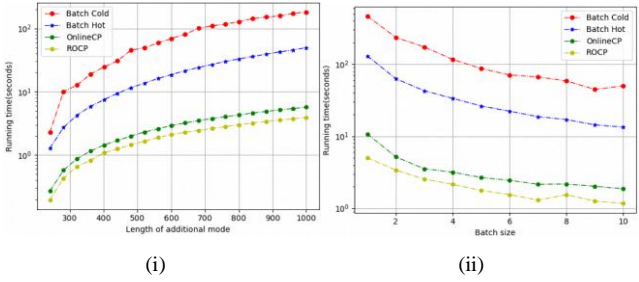

Fig. 2. The scalability with different batch sizes. (i) Running time of adding a slice at time t. (ii)Running time of different batch sizes

From Fig. 2 (i), we observe that ROCP algorithm is the fastest no matter how big the last dimension of tensor is. From Fig. 2 (ii), we could safely make a conclusion that ROCP algorithm is the fastest in all batch size range. In addition, we can find that the running time will become shorter when the batch size becomes bigger.

*E. Sensitivity to Initialization*

In order to explore the impact of initialization, a tensor whose size is 100×100×100 is used in the experiment and each method run 10 trails on this tensor to get average results. The results are reported in Table V.

TABLE V. RESULTS ARE DISPLAYED AS MEAN±STD, WHERE MEAN IS AVERAGE FINAL FITNESS AND STD IS THE STANDARD DEVIATION.

| Algorithm | Fitness (%) |
|---|---|
| Batch Cold | 90.78 |
| Batch Hot | 79.25±15.95 |
| Online CP | 76.51±16.90 |
| ROCP | **86.95±7.11** |

From the Table V, we can see that the fitness of OnlineCP is lower than that of ROCP. The reason is that when using the offline algorithm to calculate the initial solution, OnlineCP often fluctuates greatly and leads to a poor final result. It indicates that OnlineCP is sensitive to initial solution. As for ROCP, its initial solution is stable. After running for several trials, the average fitness of ROCP is much better than that of OnlineCP.

## V. CONCLUSION

This paper presents randomized online CP decomposition method for processing large-scale tensor. The theoretical analysis and experimental results show that the processing time and memory usage have been reduced by ROCP, especially for large-scale tensor. ROCP algorithm can be widely used for the scenarios highly demanding on processing speed or even in real-time processing.

In the future work, we will explore random sampling in other tensor decompositions such as Tucker decomposition, online Tucker decomposition or functional tensor decomposition [19]. Another potential direction is the combination of ROCP and other learning methods, such as kernel learning, tensor learning machines, to solve practical problems.

ACKNOWLEDGMENT

This work is partially supported by the National Natural Science Foundation of China (Grant No.61273295, 61502175 and 11501219), the Guangdong Natural Science Funds (2015A030310298), the Science and Technology Program of Guangzhou (201607010069).